\documentclass[mlabstract]{jmlr}

\usepackage{booktabs}
\usepackage{placeins}

\jmlrvolume{}
\firstpageno{1}
\jmlryear{2026}
\jmlrworkshop{}
\editors{}

\jmlrproceedings{}{}
\jmlrpages{}
\jmlryear{}

\title[More Data Cannot Break a Symmetry]{More Data Cannot Break a Symmetry:\\
Identifiability by Design}

\author{\Name{Jing Xu} \Email{jxu101@ur.rochester.edu}\and
  \Name{Christopher Kanan} \Email{ckanan@cs.rochester.edu}\\
  \addr Department of Computer Science, University of Rochester}

\begin{document}
\maketitle

\begin{abstract}
Unsupervised representational alignment recovers a stimulus-by-stimulus
correspondence from geometry alone, but the automorphism group of the stimulus
geometry bounds what any such alignment can identify, before data exist. The
obvious diagnostic for this degeneracy, the cheapest non-identity relabelling,
ranks two published designs in the wrong order, because dense sampling creates
near-duplicates whose transposition is nearly free. We turn this known invariance \citep{demetci2024breaking} into a design-time
diagnostic and intervention. In colour, where candidate geometries have closed
form, we show that the failure is structural:
sixty-four times the restart budget leaves a symmetric design unmoved while an
asymmetric set at the same $N$ recovers every time. Discriminating
representational models and recovering a correspondence are essentially
uncorrelated objectives ($r = -0.02$ over $3{,}000$ subsets). Choosing nine
colours by this diagnostic alone, without consulting any learned
representation, moves
all 93 model representations away from the degenerate point and cuts
catastrophic alignment failures from $75\%$ to $2\%$ with the models, the
layers, $N$ and the solver all held fixed. The same risk arises wherever a
regular design meets its candidate geometry's isometry group, evenly spaced
orientations, tones, or motion directions, and the check costs one function
call before data collection.
\end{abstract}

\section{Introduction}
\label{sec:intro}
Comparing representational geometries across observers, species, or between
brains and models is a central problem in neural representation
\citep{kriegeskorte2008rsa, williams2021shape}. When the stimulus
correspondence is itself the question, unsupervised alignment recovers it from
geometry alone, minimising a Gromov--Wasserstein objective over permutations
\citep{memoli2011gromov}.

The recovered correspondence inherits the symmetries of the stimulus geometry.
\citet{demetci2024breaking} show that isometric ties make the
Gromov--Wasserstein solution non-unique and propose feature priors to break
them; \citet{pashakhanloo2026stimulus} show independently that stimulus
symmetries confound representational similarity analyses. We ask the question
one step earlier: the automorphism group of the stimulus geometry is a property
of the experimental design, so identifiability is an experimental design
variable: one the experimenter controls before any data exist.

The known invariance has non-obvious consequences for experimental design.
We demonstrate them in colour, where candidate geometries have closed form
and the symmetry is exact. Our contributions are:
(i)~the obvious diagnostic ranks published designs in the wrong order
(Section~\ref{sec:formal}); (ii)~the failure is structural, not computational
(Section~\ref{sec:recovery}); (iii)~model discrimination and correspondence
recovery are essentially uncorrelated (Section~\ref{sec:design}); and
(iv)~a design-time intervention chosen without consulting learned
representations cuts catastrophic alignment failures from $75\%$ to $2\%$
across 93 held-out model representations.

\section{Identifiability of an unlabelled correspondence}
\label{sec:formal}

Two systems respond to the same $N$ stimuli, giving dissimilarity matrices
$D^1$ and $D^2$. Unsupervised alignment recovers a permutation $\pi \in S_N$
minimising $\mathcal{L}(\pi) = \sum_{i,j} ( D^1_{ij} - D^2_{\pi(i)\pi(j)}
)^2$ \citep{memoli2011gromov}, comparing structure rather than coordinates.

\begin{definition}[Automorphism group of a geometry]
\label{def:aut}
$\operatorname{Aut}(D) = \{\sigma \in S_N : D_{\sigma(i)\sigma(j)} = D_{ij}
\text{ for all } i,j\}$, the relabellings that leave every pairwise
dissimilarity unchanged. It is a subgroup of $S_N$.
\end{definition}

\begin{theorem}
\label{thm:orbit}
For every $\sigma \in \operatorname{Aut}(D^1)$ and every $\pi$,
$\mathcal{L}(\pi \circ \sigma) = \mathcal{L}(\pi)$. The correspondence is
therefore identifiable at best up to an orbit of $\operatorname{Aut}(D^1)$,
irrespective of sample size or optimiser.
\end{theorem}

\begin{figure}[t]
\centering
\includegraphics[width=\textwidth]{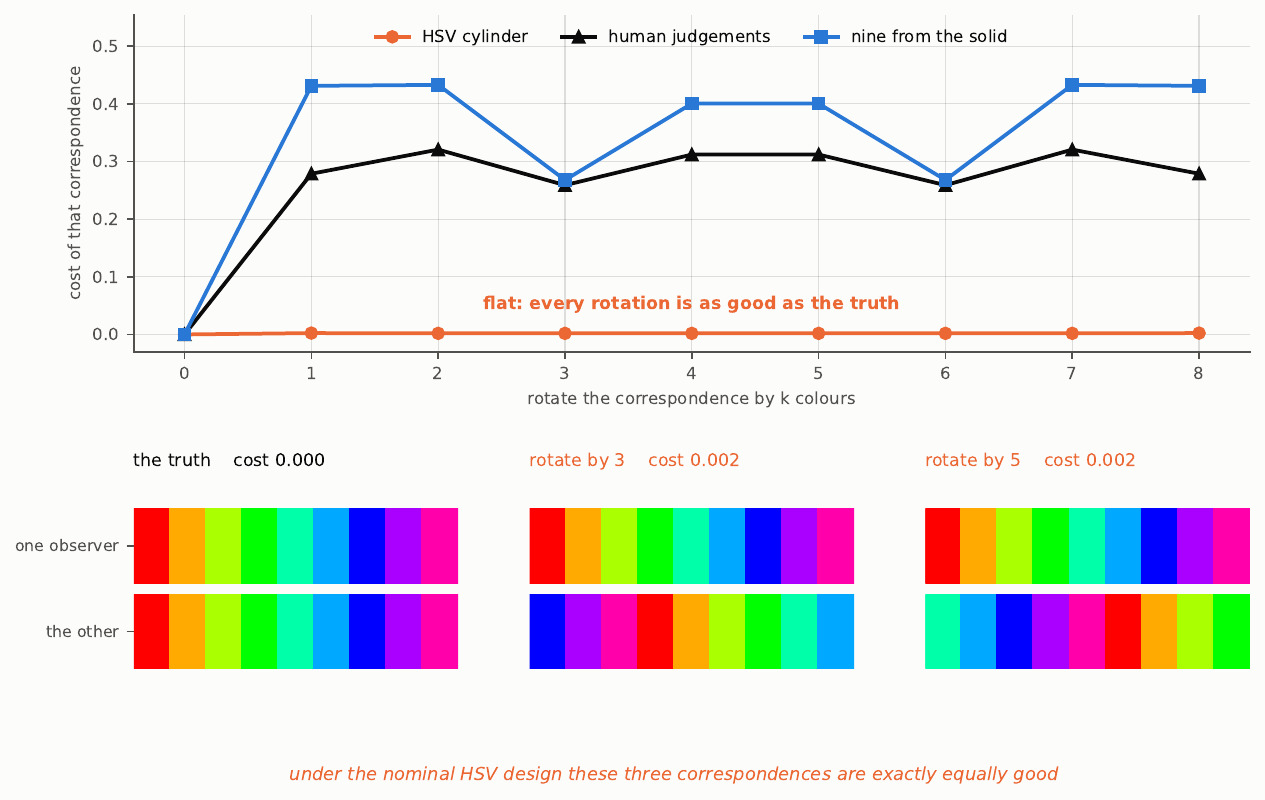}
\caption{Under the HSV cylinder every rotation of the nine-colour set costs
within $0.003$ of the identity (the nominal design is exactly equally spaced;
using the calibrated sRGB values supplied with the dataset introduces
deviations of at most $0.003$); under the observers'
own judgements (rotation cost $0.26$) or under nine colours drawn from the
colour solid, it does not.
Lower panels: the two correspondences a zero cost cannot distinguish from the
truth.}
\label{fig:rotation}
\end{figure}

The proof is a direct substitution and reindexing
(Appendix~\ref{apd:robust}). The invariance is known: \citet{demetci2024breaking} discuss the isometric ties
it produces and add feature priors to break them once the data are in hand.
On equally spaced stimuli, however, any rotation-equivariant feature prior
inherits the same group and therefore cannot break the tie. Our question comes first: $\operatorname{Aut}(D)$ depends on the stimulus set and the
candidate geometry, and the experimenter controls the first. Nine hues equally spaced at
fixed saturation and value are the case at issue: under the HSV cylinder, a
constant added to hue is a rigid rotation, so $\operatorname{Aut}(D) \supseteq
D_9$ and the objective is flat across all nine rotations
(Figure~\ref{fig:rotation}). A set invariant under a group acting by isometries
inherits that group regardless of set size, so collecting more stimuli along
the invariant submanifold cannot help.

\paragraph{The obvious diagnostic is backwards.} The natural measurement is the
cheapest non-identity relabelling, $\min_{\sigma \neq e} \lVert D - P_\sigma D
P_\sigma^{\top} \rVert_F / \lVert D \rVert_F$, and it ranks the two published
colour sets the wrong way round: enumerating all $9!$ relabellings of the
nine-colour set \citep{hirao2025fmri} gives $0.0794$ under CIELAB, while the
93-colour set that supports a published alignment \citep{kawakita2025red} gives
$0.0025$. Dense sampling creates near-duplicates whose transposition is nearly
free, so this minimum gets easier to beat the more one collects.

What matters is not how cheap a wrong answer is but how wrong a cheap answer
can be. We therefore define the \emph{catastrophic cost} as the cheapest
relabelling that renames at least half the stimuli, counting a stimulus as
renamed when it moves more than a quarter of the set's diameter:
a relabelling this large destroys any stimulus-level inference the alignment
was meant to support.
The nine-colour set is then exactly $0$ and the 93-colour set is $0.33$.
The same zero appears wherever a regular design meets an isometric group:
twelve orientations, twelve motion directions, twelve tones within an octave,
and a $5\times5$ factorial grid all have catastrophic cost exactly zero
(Table~\ref{tab:designs} in Appendix~\ref{apd:robust}). Shown twelve evenly
spaced gratings, learned representations come within $0.0001$ of that zero
(Appendix~\ref{apd:orientation}).

\section{Structural failure, not optimisation failure}
\label{sec:recovery}

\begin{figure}[t]
\centering
\includegraphics[width=\textwidth]{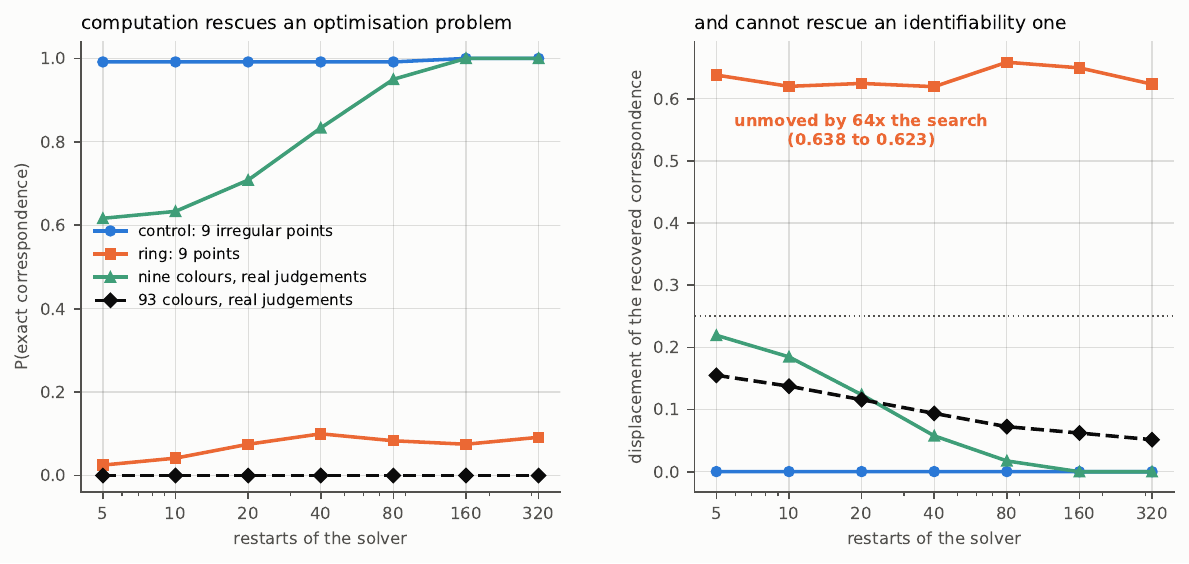}
\caption{Computation rescues an optimisation problem and cannot rescue an
identifiability one. The ring is unmoved by sixty-four times the search;
the 93-colour displacement falls to $0.052$ while exact accuracy stays at
$0\%$, which is why displacement is the primary outcome.}
\label{fig:restarts}
\end{figure}

Any claim that an alignment failed invites the reply that the optimiser was not
run hard enough \citep{takeda2025toolbox}. The distinction is settled by
experiment. Holding the data fixed and varying only the restart budget from 5
to 320 (Figure~\ref{fig:restarts}), the real nine-colour judgements climb from
$62\%$ to $100\%$ exact recovery and the 93-colour displacement falls from
$0.155$ to $0.052$, while the synthetic ring does not move: $0.638$ to $0.623$
with sixty-four times the search. Nine irregular points with same $N$, same
noise, same protocol recover at every budget and never catastrophically, so this
failure is a property of the geometry.\footnote{We use the exact
conditional-gradient solver on normalised matrices; entropic regularisation has
no usable setting at this $N$ (Appendix~\ref{apd:robust}).}

On a degenerate design the solver returns the orbit itself: across 300
repetitions every solution on the ring is an exact dihedral element at cost
$0.0000$, while on real judgements it is the identity every time.

Exact matching accuracy is the wrong primary outcome: on the 93-colour data it
is $0\%$ at every budget because some near-duplicate pair is always swapped,
while displacement falls to $0.052$ and no run is catastrophic.

That the real judgements recover at all is the honest middle of this paper.
Observers are not the HSV cylinder: their hue asymmetry costs a rotation
$0.26$ where the cylinder costs $0.003$, so the design's own degeneracy is
never exercised. The design is rescued by the data, not by the design itself.

\section{Designing a set that can answer the question}
\label{sec:design}
If the symmetry is the problem, only breaking it helps. Sampling more stimuli
along the same invariant submanifold, with more hues at $S = V = 1$, leaves the
catastrophic cost at exactly zero from $N = 6$ to $N = 72$. Varying saturation
and value while keeping the same hue spacing lifts the cost immediately: at
$N = 9$ the matched-$N$ contrast is $0.000$ against $0.188$ and $0.231$
(Figure~\ref{fig:curve} in Appendix~\ref{apd:robust}). The published 93-colour
set falls on the same curve at its own $N$: more data alone does not explain
the improvement, breaking the invariant structure does. The lever is saturation
and value, not hue uniformity (Rayleigh $p = 0.136$).

Breaking the symmetry might cost model-discriminating power, forcing a
trade-off. We find no systematic trade-off: across $3{,}000$ random
nine-colour subsets,
discriminability \citep{golan2022controversial} and catastrophic cost are
essentially uncorrelated ($r = -0.02$), and none of the $3{,}000$ is
degenerate: the published set is the only zero.

The prescription is two lines (Figure~\ref{fig:method}). Before collecting data,
compute the catastrophic cost under every candidate geometry and take the worst
case. If it is zero, vary the sampling away from the invariant submanifold;
most gains saturate around $N \approx 18$--$24$ in
our simulations. The prescription is not merely precautionary: changing only which nine colours
are shown takes catastrophic alignment failures across 93 model representations
from $75\%$ to $2\%$ (Appendix~\ref{apd:models}).


\bibliography{refs}

\appendix
\section{Robustness of the diagnostic}
\label{apd:robust}

\begin{table}[t]
\centering
\caption{Catastrophic cost across subfields. Zeros marked exact are achieved
by the design's own symmetry group and so are not search bounds. Each is
conditional on its candidate geometry exactly as the colour result is: the
orientation zero assumes orientation is an angle on a circle, the standard
model rather than a fact. THINGS uses human similarity embeddings from the SPoSE
model \citep{hebart2020things}.}
\label{tab:designs}
\begin{tabular}{llll}
\toprule
Design & Field & Group & Catastrophic cost \\
\midrule
Nine saturated hues            & colour        & $D_9$    & $0.000$ (exact) \\
Twelve orientations            & V1            & $D_{12}$ & $0.000$ (exact) \\
Twelve motion directions       & MT            & $D_{12}$ & $0.000$ (exact) \\
Twelve tones within an octave  & audition      & $D_{12}$ & $0.000$ (exact) \\
$5\times5$ factorial grid      & psychophysics & $D_4$    & $0.000$ (exact) \\
THINGS objects ($n=200$)       & object vision & ---      & $0.224$ \\
\bottomrule
\end{tabular}
\end{table}

\paragraph{Proof of Theorem~\ref{thm:orbit}.} Substituting $k = \sigma(i)$,
$l = \sigma(j)$ and reindexing the sum, $\mathcal{L}(\pi\sigma) = \sum_{k,l}
( D^1_{\sigma^{-1}(k)\sigma^{-1}(l)} - D^2_{\pi(k)\pi(l)} )^2$. Since
$\operatorname{Aut}(D^1)$ is a group, $\sigma^{-1}$ lies in it and
$D^1_{\sigma^{-1}(k)\sigma^{-1}(l)} = D^1_{kl}$.

\paragraph{Catastrophic cost thresholds.} Counting stimuli rather than
averaging displacement keeps the measure comparable across set sizes: a
transposition moves at most $2/N$ of the set, so a threshold on the mean means
something different at every $N$.

\paragraph{Entropic regularisation.} Entropic Gromov--Wasserstein replaces the
hard permutation with a soft coupling weighted by $\varepsilon$. At $N = 9$
there is no usable setting: below $\varepsilon \approx 10^{-3}$ the coupling
underflows to zero while the log reports a distance of zero, which a
lowest-cost multistart then actively selects; above it the coupling is exactly
uniform. All results in the main text use the exact conditional-gradient solver
on normalised matrices.

The catastrophic cost is robust to the choice of displacement threshold
$\tau$ and minimum-moved fraction: across a $5 \times 3$ grid
($\tau \in \{0.15, 0.20, 0.25, 0.30, 0.40\}$, minimum fraction
$\in \{0.3, 0.5, 0.7\}$), the nine-colour hue ring scores zero in all 15
cells, the published nine-colour set scores zero in 14 of 15, and the
93-colour set is never below $0.31$.

The zero depends entirely on the candidate geometry. Removing the HSV
cylinder from the candidate set lifts the worst case to $0.040$
(opponent alone: $0.040$; CIELAB alone: $0.159$). The correct framing is
that the experimenter does not know the geometry at design time, so the
worst case is the right criterion, even though the measured human geometry
turned out not to exercise it (Section~\ref{sec:recovery}).

The risk is specific to designs an experimenter builds to be regular. A set
carries an exact automorphism group only when it is sampled invariantly under a
group acting by isometries on the candidate geometry, which is what evenly
spaced hues, orientations, motion directions and factorial grids all do by
construction. An irregular semantic set need not show that exact degeneracy:
200 objects drawn from the THINGS similarity embedding score $0.224$,
comparable to the non-degenerate colour designs and far from zero
(Table~\ref{tab:designs}). The diagnostic can therefore screen a natural or
semantic stimulus set for detectable degeneracy as well as reject a regular
one, though a nonzero search result does not prove that none exists, and we
make no claim that representational alignment is at risk in general.

Nothing in the argument is about circles, about colour, or about dimension.
What makes a design degenerate is one automorphism that moves a large part of
the set, and any orbit of a group acting by isometries supplies one as soon as
some group element displaces at least half the stimuli. The twelve vertices of
an icosahedron in three dimensions, the sixteen of a hypercube in four, the
thirty-two of a five-cube, a cross-polytope in four and a regular simplex in
six all have catastrophic cost exactly zero, while matched-size random point
sets in the same spaces score between $0.31$ and $0.36$. The icosahedron needs
only the antipodal map: a large symmetry group is not required, one element
with a large motion is.

This is also where the measure has to be used carefully. It is a minimum over
a pool of candidate relabellings, so a zero is a proof and a nonzero value is
only an upper bound, and a pool of random relabellings will not find the
antipodal map of an icosahedron among the $12!$ possibilities. The design's own
symmetry group has to be handed to the check, which is what our implementation
does for the dihedral case and what any use of it on a new design requires.

\begin{figure}[t]
\centering
\includegraphics[width=\textwidth]{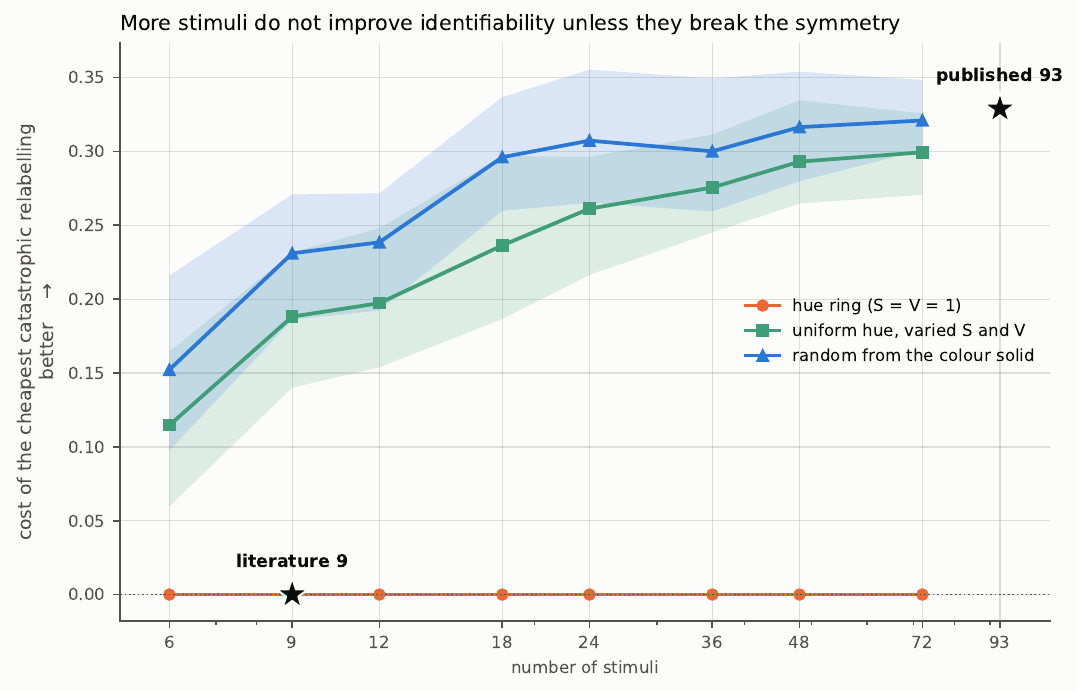}
\caption{Catastrophic cost against set size for three sampling schemes. Keeping
stimuli on the hue circle at $S = V = 1$ (orange) leaves the cost at zero
regardless of $N$; any departure from the invariant submanifold lifts it
immediately. The published 93-colour set (star, upper right) falls on the curve
at its own $N$: it is not special, just not on the circle. The matched-$N$
contrast at $N = 9$ rules out the reading that the 93-colour success is simply
more data. Bands show 10th--90th percentile over 40 random draws.}
\label{fig:curve}
\end{figure}

\begin{figure}[t]
\centering
\includegraphics[width=0.85\textwidth]{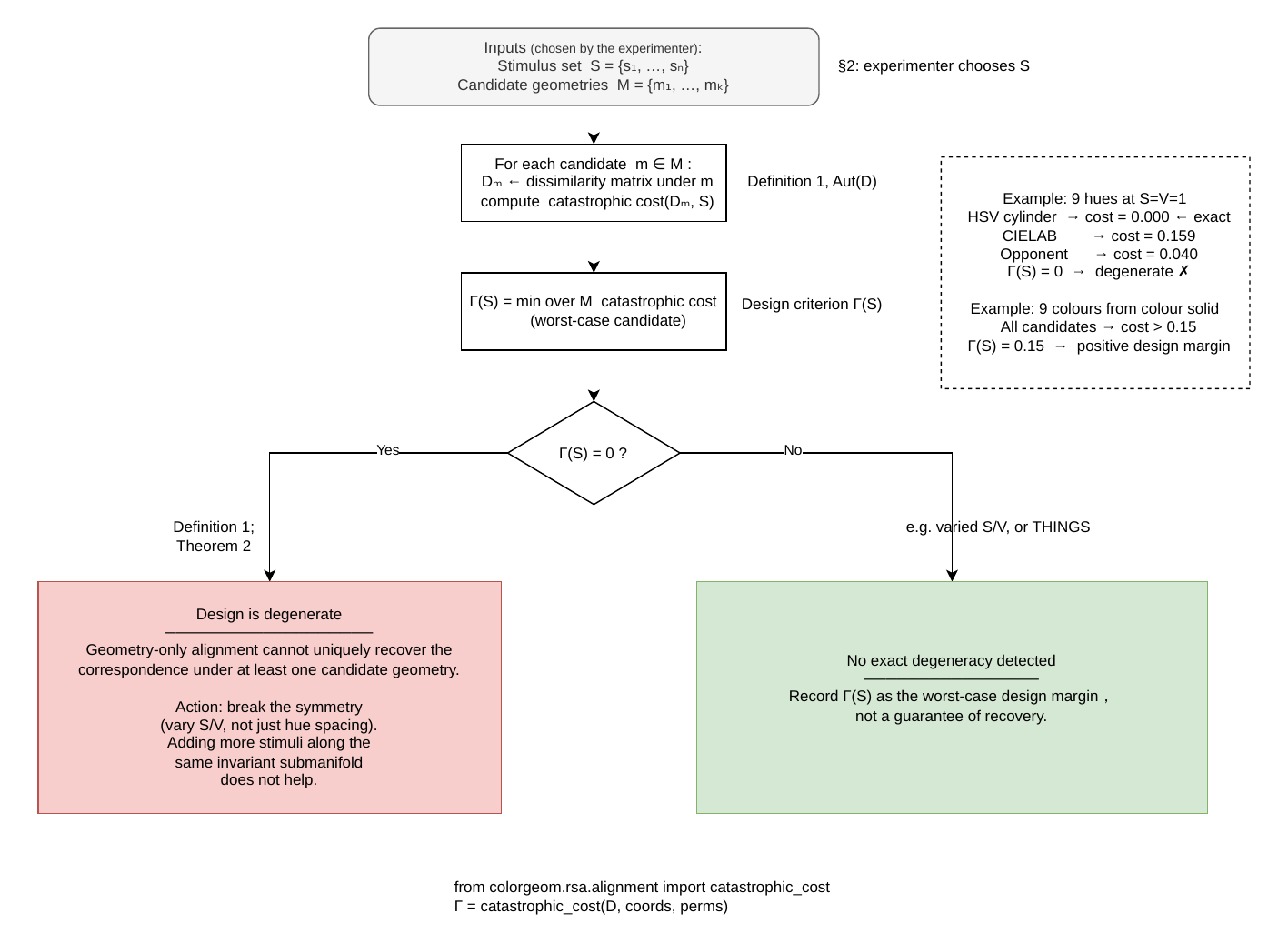}
\caption{Design-time prescription. Before data collection, compute the
catastrophic cost of the stimulus set under every candidate geometry and take
the worst case $\Gamma(S)$. A zero means the correspondence cannot be
identified under that geometry; the remedy is to vary the sampling away from the
invariant submanifold, not to add more stimuli along it.}
\label{fig:method}
\end{figure}

\FloatBarrier
\section{Model representations}
\label{apd:models}

\begin{figure}[t]
\centering
\includegraphics[width=\textwidth]{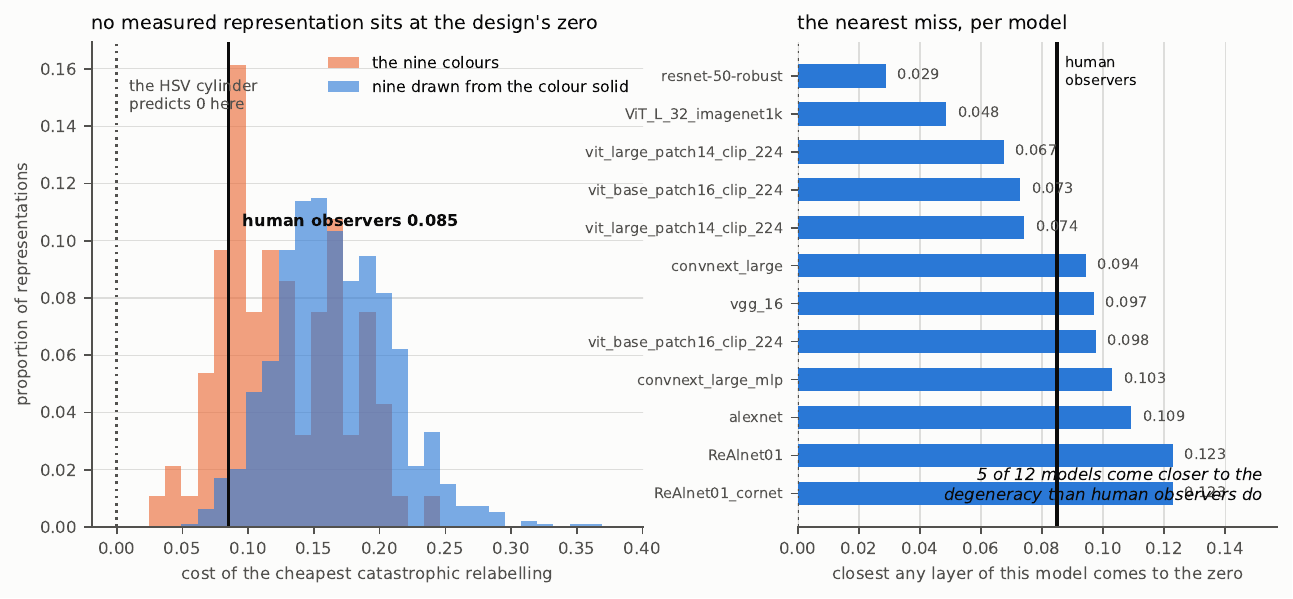}
\caption{Catastrophic cost across 93 model representations. Left: the published
nine-colour set against 930 matched-size draws from the colour solid, the same
representations and the same $N$ in both cases. No representation sits at the
degenerate point, and the solid draws sit further from it. Right: the closest
any layer of each model comes to zero. The line marks the human observers'
catastrophic cost ($0.085$); five models have at least one layer below it.}
\label{fig:models}
\end{figure}

We compute the catastrophic cost of the nine-colour set under 93
representations extracted from 12 vision models spanning convolutional
(AlexNet, VGG-16, ResNet-50-robust, ConvNeXt-Large, ConvNeXt-Large-MLP,
ReAlnet, CORnet), transformer (ViT-L/32), and multimodal (CLIP ViT-Base and
ViT-Large, four pretraining datasets) architectures
(Figure~\ref{fig:models}). The median catastrophic cost is $0.123$ and the
minimum is $0.029$ (adversarially trained ResNet-50); no representation falls
below $0.01$. Five of twelve models have at least one layer closer to the
degenerate point than the human observers ($0.085$): the stimulus set's exact
$D_9$ symmetry under HSV does not transfer to any learned representation we
tested, but some come closer than the observers do.

The comparison that isolates the design is against matched-size draws from the
colour solid, where the representation, the layer and the number of stimuli are
all held fixed and only the stimulus set differs. Across 930 such draws the
median catastrophic cost is $0.163$ against $0.123$ for the published ring, the
minimum is $0.058$ against $0.029$, and no draw falls below $0.05$ where
$3.2\%$ of the ring representations do. The advantage the diagnostic predicts
from the candidate geometries is therefore also present in the learned
representations themselves.

Whether the diagnostic predicts anything about alignment itself is a separate
question, and we test it by intervention. Four designs matched at nine stimuli
are shown to the same 93 representations: the published hue ring, a set
optimised to maximise catastrophic cost, a set optimised to separate the three
candidate geometries, and a random draw from the colour solid. The three
alternatives are drawn from the dense grid that was rendered and passed through
the models alongside the published set, so every design is a lookup into
features that already exist and no representation is recomputed.
The optimised sets are chosen from the candidate geometries alone, before any
model representation is consulted, so the 93 representations are unseen at
design time. Unsupervised alignment then runs on split halves of each
representation's units, where the identity is the correct answer and there is
almost no measurement noise to blame. Holding the models, the layers, $N$ and
the solver fixed and changing only which nine colours are shown takes exact
recovery from $8.6\%$ to $52.7\%$, mean displacement from $0.415$ to $0.034$,
and the catastrophic failure rate from $75.3\%$ to $2.2\%$. The shift is not
carried by a subset: all 93 representations move away from the degenerate point
(median $+0.207$, bootstrap CI $[0.199, 0.218]$, Wilcoxon $p < 10^{-16}$), and
68 representations cross from a catastrophic correspondence to a
non-catastrophic one while none crosses the other way.

\begin{figure}[t]
\centering
\includegraphics[width=\textwidth]{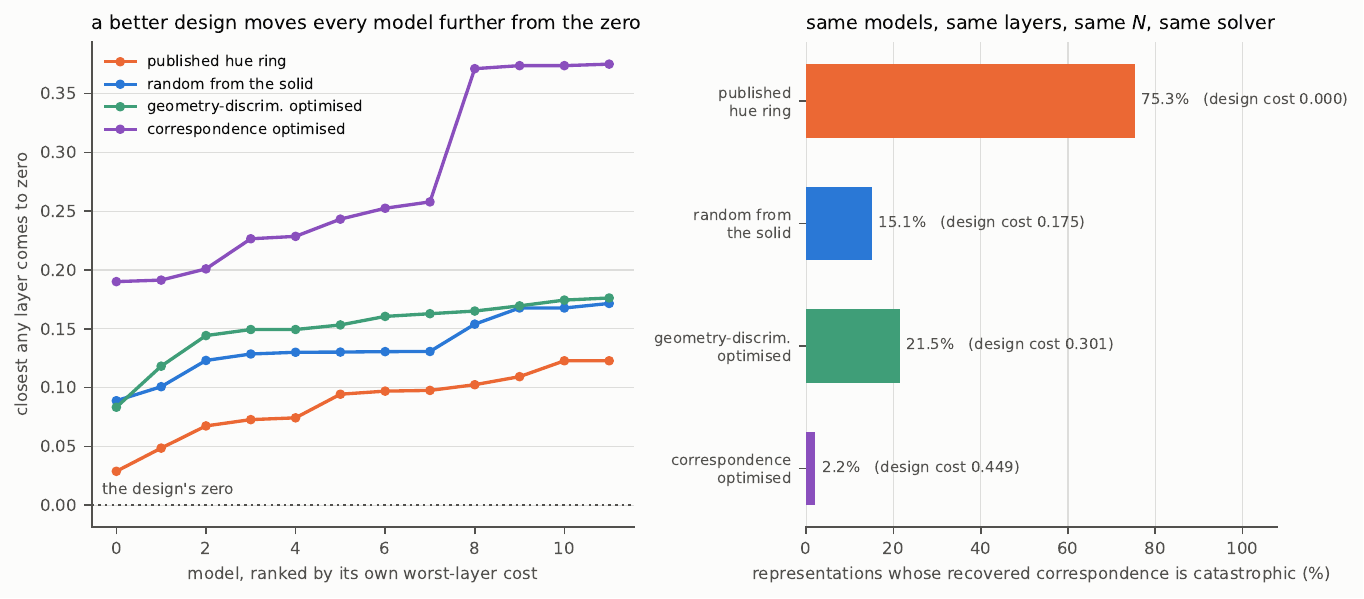}
\caption{Four designs matched at nine stimuli, shown to the same 93
representations. Left: the closest any layer of each model comes to zero,
models ranked within each design. The correspondence-optimised set moves every
model further out, though the spread grows in proportion to the shift rather
than beyond it (SD over median $0.30$ against $0.29$ for the ring), so the
models are not made more separable. The geometry-discrimination set tracks the
random draw almost exactly. Right: how often the recovered correspondence is
catastrophic, with the design-time cost beside each bar.}
\label{fig:intervention}
\end{figure}

Their downstream effects also dissociate. The set optimised to tell the candidate geometries apart is no better
for correspondence than a random draw from the solid (mean displacement $0.163$
against $0.164$, $p = 0.99$), while the correspondence-optimised set beats that
same random draw by a wide margin ($p < 10^{-10}$). Choosing stimuli that
discriminate representational hypotheses does not on its own produce stimuli
that identify a correspondence (Figure~\ref{fig:intervention}).

Leaving the zero is not the same as being safe. A random draw from the colour
solid scores $0.175$, the same order as the $0.224$ of the THINGS objects and
as the other broad-support designs tested here, and it still leaves $15.1\%$ of
the representations with a catastrophic correspondence; reaching $2.2\%$ takes
a design chosen for that purpose. Design cost does not order the outcomes
exactly, and we do not claim it calibrates them: the geometry-discrimination
set scores higher than the random draw and does slightly worse.

In this experiment the correspondence-optimised design does not also separate
the representations. Comparing the nine-stimulus dissimilarity matrices of the 93
representations directly, as one minus the rank correlation between them, the
median distance between two representations falls from $0.555$ under the
published ring to $0.101$ under the correspondence-optimised set: the better
design makes the models agree, not differ. The ring makes them look most
unlike each other, but that apparent difference is not usable, because split
halves of a single representation already fail to agree on the ring three times
in four. Much of the ring's apparent model separation therefore appears to
reflect within-representation instability rather than stable between-model
structure.

\FloatBarrier
\section{Orientation}
\label{apd:orientation}

Colour is a favourable case, so the same analysis is run on a different visual
variable with a different standard model. A grating's orientation is
$\pi$-periodic and is embedded as $(\cos 2\theta, \sin 2\theta)$, under which
twelve evenly spaced orientations are a regular twelve-gon carrying $D_{12}$
exactly. We render twelve evenly spaced gratings, the same twelve jittered, and
twelve unevenly spaced ones, identical in spatial frequency, phase, contrast and
aperture, and pass all three through the twelve models that produced the colour
results.

The degeneracy is more severe here than in colour. Under the evenly spaced
design the closest representation comes within $0.0001$ of the exact degenerate
point and half of the 93 representations score below $0.05$, against a minimum
of $0.029$ and no representation below $0.01$ for the nine-colour ring. Learned
representations break the colour design's symmetry but preserve the grating
design's much more closely. Both alternatives lift nearly every representation
away from it (median
$+0.103$ for the uneven set, 92 of 93, $p < 10^{-16}$), and split-half
alignment improves accordingly, from $51.6\%$ catastrophic under the regular
design to $19.4\%$ under the uneven one (61 of 93 improved, $p < 10^{-6}$, 36
representations fixed against 6 broken).

Jitter alone is not a remedy. Perturbing the evenly spaced orientations raises
the design-time cost reliably (92 of 93 representations, $p < 10^{-16}$) and
leaves alignment where it was ($p = 0.51$, 16 fixed against 12 broken). Here a
small jitter of the regular design behaves like the regular one, which is the reason
the prescription asks for sampling away from the invariant submanifold rather
than for a small perturbation of it.

\FloatBarrier
\section{fMRI}
\label{apd:fmri}

We analyse the fMRI data of \citet{hirao2025fmri}: all 35 recruited subjects
performing sequential colour similarity judgements at 3T (the original study
excluded four for head motion; we analyse all 35 after independent
fMRIPrep-based preprocessing and did not run a separate motion-exclusion step,
so we cannot rule out that those four influence the estimates). Regions of
interest are V1, hV4 and hV4+VO1 from the \citet{wang2015probabilistic}
probabilistic atlas; no individual retinotopic mapping is available.

A permutation-based ceiling test (1000 permutations of the crossnobis
dissimilarity matrix) asks whether cross-subject structure exceeds the noise
floor. In hV4 alone under the no-report condition, the lower bound on the
noise ceiling is $+0.127$ ($p = 0.004$); across six analysis variants (ROI
definition $\times$ trial selection), lower bounds range from $0.104$ to
$0.127$ ($p = 0.004$--$0.017$), all surviving Benjamini--Hochberg correction
at $q = 0.05$ across 17 tests. V1 does not reach significance under the same
test in the no-report condition ($+0.061$, $p = 0.083$). The upper bound of the
noise ceiling is
$0.21$--$0.22$, compared with $0.89$ for behavioural judgements. Colour
geometry in hV4 is reliably present across observers but substantially noisier
than the behavioural geometry used for alignment in the main text (upper bound
$0.21$--$0.22$ versus $0.89$ for behaviour).

Several limitations constrain interpretation: the ROIs are atlas-based, not
individually localised; no fieldmaps are available for distortion correction;
stimuli are presented for 300\,ms in a rapid event-related design; and no
functional localiser is run. The distinction between ``no shared structure''
and ``structure too noisy to align'' cannot be resolved at this noise ceiling.

\FloatBarrier
\section{Four measurements that cannot separate the candidates}
\label{apd:motivation}

On the nine-colour set at $S = V = 1$, four independent measurements fail to
distinguish the three candidate colour geometries (HSV cylinder, CIELAB,
cone-opponent), motivating the identifiability question of
Section~\ref{sec:formal}.

Canonical correlation analysis gives pairwise correlations of
$0.972$--$0.984$ across the three geometry pairs, and the HSV representation
is rank-deficient: constant saturation and value reduce the three-dimensional
cylinder to a one-dimensional circle, collapsing one canonical dimension
entirely. Human similarity judgements correlate with the hue-circle prediction
at $\rho = 0.800$ (noise ceiling $0.890$). Linear decoding from twelve vision
models reaches $R^2 = 0.98$--$0.99$ at best for all three coordinate systems,
and even raw sRGB pixels reach $R^2 = 0.70$--$0.95$; the stimuli vary only
in colour, so any non-constant representation trivially encodes it.
Cross-subject structure in V1 and hV4 is too coarse to separate candidates
(Appendix~\ref{apd:fmri}). None of these four measurements can determine
which colour geometry a system uses; the question is whether unsupervised
alignment can, and if so, under what stimulus design.

\FloatBarrier
\section{Probe-set identifiability beyond perception}
\label{apd:probes}

Theorem~\ref{thm:orbit} is a statement about dissimilarity matrices and
permutations, not specifically about colour or perception. More generally, let
a probe set $S$ induce a dissimilarity matrix $D_m(S)$ under a candidate
representational geometry $m$. Whenever two systems are compared using only the
geometry induced by a shared set of probes, non-trivial automorphisms of
$D_m(S)$ bound the correspondence that geometry-only alignment can uniquely
recover. The colour ring provides an analytically transparent case in which
this symmetry is exact and available in closed form; an important open question
is how often approximate versions arise in less structured domains.

A natural extension is to learned representations compared across models or
training runs, where the probes may be images, prompts, or other model inputs.
A restricted probe distribution can fail to distinguish latent features that
would separate under other inputs. If two features have identical response
profiles over the entire sampled support, then collecting additional probes
from that same symmetry-preserving support cannot identify their
correspondence: more data within the invariant submanifold cannot break the
symmetry any more than more hues at fixed saturation and value can break
$D_9$. More generally, approximately redundant response profiles may produce
approximate ambiguities whose practical importance depends on their magnitude
relative to noise and on whether the sampling distribution ever reaches
inputs that separate them. The design implication parallels the colour
results: rather than increasing sample density along a single manifold, one
should add probes that vary along directions capable of breaking the relevant
ambiguity.

This probe-induced ambiguity is distinct from parameterisation symmetries such
as permutations of hidden units within a network. Parameterisation symmetry is
a property of equivalent model parameterisations and exists independently of
the chosen probes; it may be handled by quotienting out or otherwise
accounting for equivalent parameterisations. The ambiguity considered here is
conditional on the probe set and the candidate representational geometry,
$(S, m) \mapsto D_m(S)$, and is reduced by choosing inputs that break the
relevant automorphisms rather than by post-hoc alignment of parameters.

We have not tested this extension empirically. The colour and orientation
experiments establish the principle in settings where the candidate geometry,
its symmetry group and the ground-truth correspondence are all known exactly
(Appendices~\ref{apd:models}--\ref{apd:orientation}). Whether approximate
symmetries induced by restricted natural-image or language probe sets create
practically important ambiguity in cross-model feature matching remains an
open question. The diagnostic introduced here is designed to answer it, and
the natural next step is actively selecting probes to maximise correspondence
identifiability before performing representational alignment.

\end{document}